\documentclass[letterpaper, 10 pt, conference]{ieeeconf}  % Comment this line out if you need a4paper

\IEEEoverridecommandlockouts                              % This command is only needed if 
\usepackage{microtype}
\usepackage{graphicx}
\usepackage{subfigure}
\usepackage{booktabs} % for professional tables
\usepackage{xspace}
\usepackage{cite}
\usepackage{makecell}
\usepackage{multirow}
\usepackage{hyperref}
\usepackage{caption}

\usepackage{amsmath}
\usepackage{amssymb}
\usepackage{mathtools}
\usepackage{wrapfig,lipsum,booktabs}
\usepackage{epsfig}
\usepackage{nccmath}
\usepackage{adjustbox}
\let\labelindent\relax
\usepackage{enumitem}
\usepackage{ragged2e}
\usepackage{cuted}

\usepackage[capitalize,noabbrev]{cleveref}

\newcommand{\our}{\textsc{Re\textsuperscript{3}Sim}\xspace}
\usepackage[textsize=tiny]{todonotes}

\title{\LARGE \bf
\textsc{Re$^3$Sim}: Generating High-Fidelity Simulation Data via 3D-Photorealistic Real-to-Sim for Robotic Manipulation}

\author{Xiaoshen Han$^{1,2,*}$ Junqiu Yu$^{2,*}$ Minghuan Liu$^{1}$ Yilun Chen$^{2,\dagger}$ Xiaoyang Lyu$^{3}$ Yang Tian$^{2}$ Bolun Wang$^{2}$ \\ 
Weinan Zhang$^{1,\dagger}$ and Jiangmiao Pang$^{2,\dagger}$ \\ % <-this % stops a space
\small Website: \url{https://re3sim.github.io/}
\thanks{$^{1}$Shanghai Jiao Tong University,
        $^{2}$Shanghai AI Laboratory,
        $^{3}$The University of Hong Kong.
        * denotes equal contribution.
        $\dagger$ denotes corresponding authors.
}%
}

\begin{document}

\maketitle
\thispagestyle{empty}
\pagestyle{empty}

\begin{strip}
  \vspace{-26mm}
  \centering
  \includegraphics[width=\textwidth]{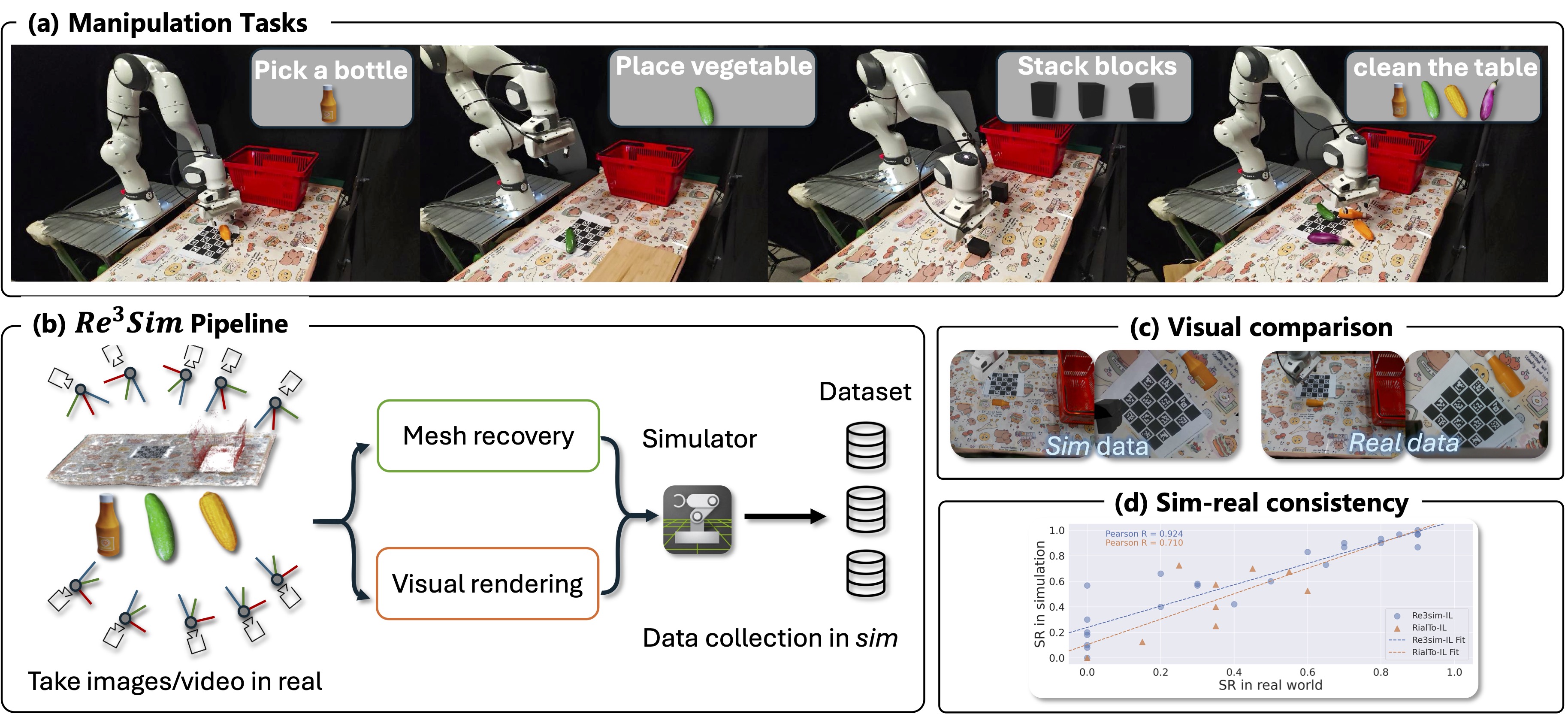}
  \captionof{figure}{\textbf{Illustration of RE$^3$SIM.} a) RE$^3$SIM allows zero-shot policy transfer on various tasks. b) The system pipeline to generate high-quality data. c) High-fidelity rendering results. d) Consistency in success rates between real and simulated environments.}
  \label{fig:teaser}
  \vspace{-0.1in}
\end{strip}

% High-quality, extensive data is the cornerstone of building powerful, robust, and generalizable models for robotics. 
% However, collecting such data in the real world normally requires designing efficient teleoperation systems, skilled operators, and costly robots, making the process resource-intensive. 
% On the contrary, simulation offers a scalable alternative for generating vast amounts of data in a short period but often suffers from sim-to-real gaps due to limitations in realistic geometric modeling and visual rendering.
\begin{abstract}
Real-world data collection for robotics is costly and resource-intensive, requiring skilled operators and expensive hardware. Simulations offer a scalable alternative but often fail to achieve sim-to-real generalization due to geometric and visual gaps. To address these challenges, we propose a 3D-photorealistic real-to-sim system, namely, \our, addressing geometric and visual sim-to-real gaps. 
\our employs advanced 3D reconstruction and rendering techniques to faithfully recreate real-world scenarios, enabling real-time rendering of simulated cross-view cameras within a physics-based simulator.
By utilizing privileged information to collect expert demonstrations efficiently in simulation, and train robot policies with imitation learning, we validate the effectiveness of the real-to-sim-to-real system across various manipulation task scenarios. 
Notably, with only simulated data, we can achieve zero-shot sim-to-real transfer with an average success rate exceeding 58\%.
To push the limit of real-to-sim, we further generate a large-scale simulation dataset, demonstrating how a robust policy can be built from simulation data that generalizes across various objects. 
% Codes and demos are available at: \url{https://anonymous-robot-ai.github.io/re3sim/}.
\end{abstract}
\vspace{-3mm}

\section{Introduction}
\label{sec:intro}
\vspace{-2mm}
We have witnessed impressive generalization capabilities of robotic models~\cite{rt2, gr2,openvla} in certain tabletop or kitchen scenarios, achieved through high-quality teleoperation data. However, collecting real-world expert data~\cite{pi0, brohan2022rt} remains a time-consuming and costly process. Despite advancements in teleoperation systems~\cite{cheng2024tv,aldaco2024aloha,yang2024ace}, the labor involved is still intensive, making this a key challenge in robotics research. Recent efforts~\cite{oxe, droid} have explored inter-institutional collaboration to address this issue. In contrast, simulation data offers the advantage of exponential scalability with computational resources, making it an appealing and renewable alternative for training robotic policies. Research on sim-to-real transfer~\cite{mandlekar2023mimicgen,jiang2024dexmimicen} has increasingly focused on generating high-quality simulated or synthetic data to train real-world robotic policies. Unfortunately, simulation data often exhibits significant sim-to-real gaps, making it necessary to collect target domain data for effective policy fine-tuning.

In real-to-sim-real scenarios, consistency with the real world in appearance and geometry is essential. 
High-quality RGB rendering ensures visual consistency between real and simulated domains, while accurate mesh reconstruction captures scene geometry for realistic interactions. 
To achieve this, we propose a 3D-photorealistic real-to-sim-to-real system, \our, short for reconstruction-rendering-based real-to-sim. 
\our faithfully replicates real-world scenarios by integrating photorealistic RGB rendering with precise 3D geometry reconstruction. 
The visual rendering is realized via Gaussian rasterization~\cite{kerbl3Dgaussians}, while the 3D geometry is reconstructed using multi-view stereo (MVS) techniques~\cite{schoenberger2016sfm, openmvs2020}. 
This dual-stage pipeline enables efficient, high-fidelity simulation, where physical dynamics are handled by physics engine backends~\cite{isaacsim,Xiang-2020-SAPIEN,todorov2012mujoco}, and real-time rendering is achieved through a dedicated hybrid rendering engine.

% Specifically, \our adopts a sequential real-to-sim pipeline comprising three key steps: (a) \textit{mesh recovery} for reconstructing scene and object geometry; (b) \textit{hybrid visual rendering} for compositing foreground and background elements, and (c) \textit{real-world alignment} to synchronize world coordinates between the real and simulated environments. Human involvement is minimal in the real-to-sim process and limited to:
% (a) Placing ArUco markers in the target scenario for real-world alignment.
% (b) Capturing photos or videos of the scene and objects, including an additional capture step to align the robot base.

Specifically, \our adopts a sequential real-to-sim pipeline comprising three key steps: (a) \textit{mesh recovery} for reconstructing scene and object geometry; (b) \textit{hybrid visual rendering} for compositing foreground and background elements, and (c) \textit{real-world alignment} to synchronize world coordinates between the real and simulated environments. Human involvement is minimal in the real-to-sim process and limited to:
(a) Placing ArUco markers for real-sim alignment.
(b) Capturing images/videos of the scene and objects, including an extra step for robot base alignment.

% Once all assets and robots are imported into the simulator, \our utilizes a privileged policy to generate high-fidelity expert data. It incorporates the following key features to enhance sim-to-real transfer:

% \begin{itemize}[leftmargin=0.1in]
%     \item \textbf{High-fidelity geometry and vision}: Achieves superior reconstruction and rendering quality, reducing the sim-to-real gap in both geometry and vision.
%     \item \textbf{Rapid scene reconstruction}: Enables novel scene reconstruction in under \textit{three} minutes of manual setup.
%     \item \textbf{Efficient rendering}: Provides 24 FPS rendering for 480p images across two independent camera views.
% \end{itemize}

% After importing all assets and robots into the simulator, \our employs a privileged policy to generate high-fidelity expert demonstrations. Several key features are incorporated to promote robust sim-to-real transfer:

% \begin{itemize}[leftmargin=0.1in] 
% \item \textbf{High-fidelity geometry and vision}: Produces accurate reconstruction and photorealistic rendering, thereby minimizing discrepancies between simulated and real-world observations. 
% \item \textbf{Rapid scene reconstruction}: Enables the reconstruction of novel scenes with less than \textit{three} minutes of manual intervention. 
% \item \textbf{Efficient rendering}: Supports real-time rendering at 24 frames per second for 480p images across two independent camera perspectives. 
% \end{itemize}
After importing all assets and robots into the simulator, \our utilizes a privileged policy to collect high-fidelity expert data. It incorporates the following features to enhance sim-to-real transfer:
% \begin{itemize}[leftmargin=0.1in] 
% \item \textbf{High-fidelity geometry and vision}: Accurately reconstructs geometry and generates photorealistic views, closely matching real-world appearances to reduce sim-to-real gaps.
% \item \textbf{Rapid scene reconstruction}: Reconstructs novel scenes with under \textit{three} minutes of manual effort.
% \item \textbf{Efficient rendering}: Achieves 24 FPS at 480p across two independent camera views.
% \end{itemize}
\begin{itemize}[leftmargin=0.1in, itemsep=0pt, parsep=0pt, topsep=0pt]
\item \textbf{High-fidelity geometry and vision}: Accurately reconstructs geometry and generates photorealistic views, closely matching real-world appearances to reduce sim-to-real gaps.
\item \textbf{Rapid scene reconstruction}: Reconstructs novel scenes with under \textit{three} minutes of manual effort.
\item \textbf{Efficient rendering}: Achieves 24 FPS at 480p across two independent camera views.
\end{itemize}

% \noindent\textbf{High fidelity}: Achieves superior reconstruction and rendering quality, significantly reducing the sim-to-real gap both in collision and in vision.\\
% \noindent\textbf{Time effectiveness}: Requires less than three minutes of manual setup to reconstruct a novel scene.\\
% % for automated simulation data generation in novel scenes.
% \noindent\textbf{Computational efficiency}: Delivers 24 FPS rendering for 480p images across two independent camera views.
% \begin{itemize}
% \item \textbf{High fidelity}: Achieves superior reconstruction and rendering quality, significantly reducing the sim-to-real gap in collision and vision.
% \item \textbf{Time effectiveness}: Requires less than three minutes of manual setup to reconstruct a novel scene.
% % for automated simulation data generation in novel scenes.
% \item 
% \textbf{Computational efficiency}: Delivers 24 FPS rendering for 480p images across two independent camera views.
% \end{itemize}

To validate the effectiveness of the pipeline, we design several tabletop tasks demonstrating the generality and applicability of the real-to-sim-to-real process, as shown in Fig.~\ref{fig:teaser} The zero-shot sim-to-real policies achieve an average success rate of 58\% with only about 10 minutes of data collection in simulation, showcasing the effectiveness of our approach. We expect these experiments to highlight the potential of the real-to-sim-to-real pipeline in reducing sim-to-real gaps, paving the way for future advancements in scalable, efficient, and generalizable robotic policy.

\section{Background}
\label{sec:background_3d_recon}
\subsection{3D Reconstruction}
3D reconstruction recovers scene geometry from inputs like images, point clouds and etc. Camera poses and sparse point clouds can be obtained from structure-from-motion approaches (e.g., Colmap) and be used to reconstruct dense point clouds via either traditional methods~\cite{schoenberger2016mvs} or network-based approaches~\cite{yao2018mvsnet}. Besides, commercial software solutions, such as ARCode and PolyCam, leverage RGB-D SLAM methods to estimate camera poses and depth information, and often include some post-processing techniques (e.g., smoothing, inpainting) for enhancing the usability and quality of the reconstructed 3D models.

\subsection{Rendering}
Rendering involves generating photorealistic images from specified camera views based on either explicit or implicit 3D representations. 
Implicit methods, like Neural Radiance Fields (NeRF~\cite{mildenhall2021nerf}), represent the 3D space using a density field encoded in a neural network and render images through volumetric rendering techniques. Explicit representations, such as meshes or point clouds, employ texture mapping~\cite{blinn1976texture}. 3D Gaussian Splatting (3DGS)~\cite{kerbl3Dgaussians} uses Gaussian primitives and adopts Gaussian rasterization for high-fidelity rendering. Compared with other methods, 3DGS achieves more realistic rendering results and offers higher rendering efficiency. In detail, each Gaussian primitive is modeled as an ellipse with a full 3D covariance matrix $\Sigma$ defined in world space, centered at point $\mu$: $G(x) = \exp({-\frac{1}{2} x^T \Sigma^{-1} x}),~$
where the covariance matrix $\Sigma = R S S^T R^T~$ of a Gaussian primitive is derived from its rotation $R$ and scaling factor $S$.
Additionally, each Gaussian is assigned a color \( c_i \) and depth-ordered using the Z-buffer. Using the alpha-blending formula, the rendered color \( C \) for a view is calculated as:
\begin{equation*}
    C = \sum_{i=1}^N \prod_{j<i} \alpha_j (1 - \alpha_i) c_i~.
\end{equation*}

\section{\our}

\label{sec: sim2real}

Robot simulators~\cite{isaacsim,pybullet,todorov2012mujoco} offer high-accurate simulation results based on underlying physics engines~\cite{physx,pybullet,ode}. However, it's hard to directly obtain an accurate 3D model of the real-world scene, and the simulators are not good at rendering high-fidelity images, leaving a large gap from real-world images.
% Recent advancements in 3D reconstruction~\cite{mildenhall2021nerf,kerbl3Dgaussians,openmvs2020,yu2024gsdf,guedon2024sugar} provide high-quality rendering results and detailed 3D geometric models, promising to bridge these gaps.
We introduce \our, a 3D reconstruction-based system, to bridge the visual gap and recover realistic 3D meshes for sim-to-real transfer.

A straightforward approach is to reconstruct objects and the background together. However, using Multi-View Stereo (MVS) methods for background reconstruction often leads to suboptimal visual quality, while techniques like SuGaR ~\cite{guedon2024sugar} struggle with accurately reconstructing flat planes, causing protrusions and indentations that reduce realism in object movement. Additionally, segmenting foreground objects for manipulation is labor-intensive. 
To address these issues, we propose separately reconstructing background meshes for collision estimation and leveraging 3DGS to improve rendering realism, aligning the two. Since object rendering constitutes a small portion of the robot's visual observations, we opt not to use 3DGS to render objects.

\begin{figure*}[!htbp]
    \centering
    \includegraphics[width=.9\linewidth]{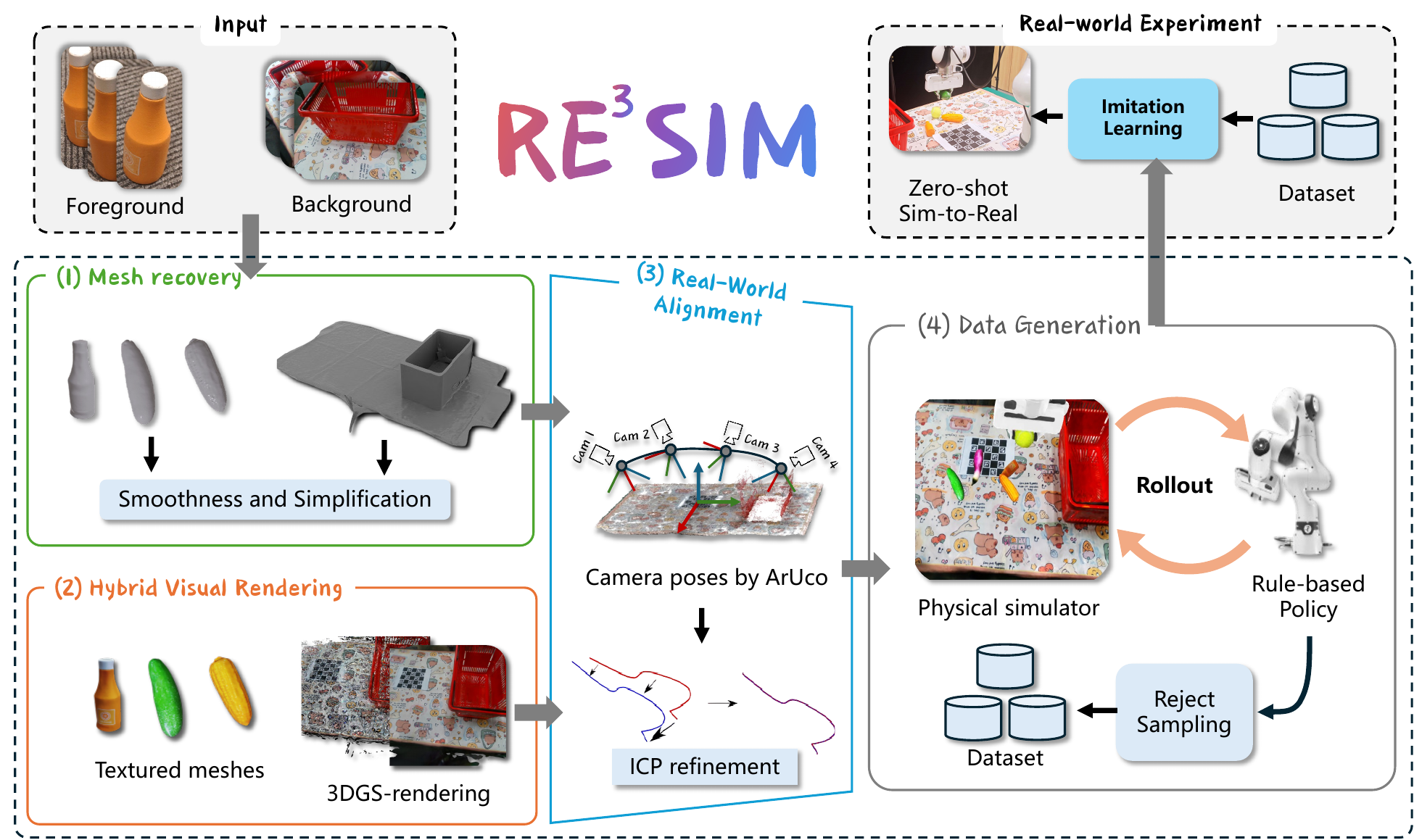}
    % \vspace{pt}
    \caption{\textbf{Illustration of the proposed real-to-sim-to-real system, \our}. It leverages 3D reconstruction and a physics-based simulator, providing small 3D gaps that enable large-scale simulation data generation for learning manipulation skills via sim-to-real transfer.}
    \vspace{-8pt}
    \label{fig:main framework}
\end{figure*}

\subsection{System Overview}
\our is a real-to-sim-real system that reconstructs meshes for collision estimation, utilizes 3D Gaussian scene representations for rendering, and generates simulation data within a physics-based simulator\footnote{Isaac Sim~\cite{isaacsim} is used by default, but other ray tracing options are also feasible.}. 
It is worth noting that we only consider reconstructing the background and foreground objects, and robots are not involved in our system since their accurate 3D assets are usually easily accessible. 
Further, we take this work as an initial study of \our, validated on a table-top robot arm for rigid body tasks.
% Subsequently, the PhysX engine within NVIDIA Isaac Sim accurately estimates the future states of robots and objects. 
% \noindent\textbf{3D reconstruction based simulation.}
As shown in Fig.~\ref{fig:main framework}, \our are composed of four components: 
mesh recovery for object collision computation, hybrid visual rendering combining 3DGS for the background and mesh-based rendering for the foreground, real-world alignment of the reconstructed scene and robot positions, and large-scale data generation for manipulation tasks.

% \begin{enumerate}[label=(\alph*)]
%     \item \textbf{Mesh recovery} (Section~\ref{sec:method:mesh_recovery}): reconstructing the mesh geometry of all objects (both foreground and background) included in the scene into the physics-based simulation for collision computation and simulation.
%     \item \textbf{Hybrid visual rendering} (Section~\ref{sec:method:rendering}): improving the visual fidelity of the scene via 3DGS. 
%     \item \textbf{Real-world alignment} (Section~\ref{sec:method:alignment}): aligning 1) the robot in the real world and the reconstructed scene, 2) the mesh and the Gaussian scene representation in simulation. 
%     \item \textbf{Data generation} (Section~\ref{sec:method:data-generation}): large-scale collection on simulation demonstrations of various manipulation skills at ease.
%     % \item \textbf{Deploy} (Section~\ref{sec:method:deployment}) ensures successful sim-to-real transfer from . 
% \end{enumerate}

\subsection{Mesh Recovery}

\label{sec:method:mesh_recovery}
This step reconstructs objects and background geometries for accurate collision detection.
% The mesh recovery step is meant for obtaining the realistic geometry information of objects in a scene, by reconstructing the mesh of the background and the objects from photos or a video, and sequentially post-processing the mesh for more accurate physics simulation. 

\noindent\textbf{Mesh reconstruction.} 
The meshes of backgrounds and objects are reconstructed separately for flexibility and are represented in USD for extensibility.
% To integrate backgrounds and objects into a physics simulator, we reconstruct their collision bodies separately. We represent them in USD to ensure high extensibility.
For the background, we first employ SfM techniques (COLMAP) to estimate the pose for each image and get the sparse point cloud based on a set of images, and then utilize OpenMVS~\cite{openmvs2020} for detailed mesh reconstruction. 
PolyCam~\cite{polycam2020} is also able to reconstruct the mesh of the background, but empirically, the geometric reconstruction results of OpenMVS exhibit fewer minor protrusions and depressions.
Subsequently, we simplify this mesh to create collision bodies. This approach effectively captures intricate details while maintaining accurate planar reconstruction. For foreground objects, we employ ARCode~\cite{arcode2022}, which can automatically segment the object, balancing usability and quality.
% providing an optimal balance between usability and reconstruction quality, ensuring efficient and precise representations.

\noindent\textbf{Post-processing.}
Reconstructed collision bodies often have voids or sharp edges due to imperfect image quality, causing erratic behavior in simulation. For example, objects may fail to rest stably on surfaces.
% or exhibit excessive bouncing when dropped from a minimal height.
Void filling and surface smoothing help improve stability and realism in simulation. 
As acquiring physical parameters like mass and friction is challenging with only initial static visual data, we use default values to simplify the process without sacrificing performance, and find it works well for many manipulation tasks.
% To mitigate such an issue, we sequentially perform void filling and surface smoothing. 
% These post-processing steps improve stability and realism in simulations. 
% Furthermore, estimating physical parameters, such as mass and friction coefficients, remains challenging when relying solely on visual data. 
% Although approaches like robot-assisted parameter acquisition~\cite{memmel2024asid} provide more accurate parameters, we found that default parameter values suffice for many manipulation tasks, simplifying the process without deteriorating the performance.

\subsection{Hybrid Visual Rendering}

\label{sec:method:rendering}

Color images, as a key type of perception signal, often yield a large visual gap between simulation and the real world.
To close such a gap, we render the foregrounds and backgrounds with hybrid real-to-sim rendering techniques.
For the foreground objects, object renderings are applied from the mesh with ray tracing support. Background reconstruction contains most parts of the scenes and applies more realistic rendering using 3DGS. Z-buffer rendering is applied to mix objects according to ground-truth depth. Notably, hybrid visual rendering supports multi-view rendering according to the perspective views. The rendering time consumption of each component is shown in \ref{tab:time_consumed}.

\subsection{Real-World Alignment}

\label{sec:method:alignment}
Despite the prior alignments between rendering and meshes, another key step is to align with real-world scenarios, specifically, the positions of the background/foreground relative to the robots. For foreground objects, we define a range of possible placements in the simulation. Regarding the background, we leverage the ArUco marker on the table to align the 3DGS coordinates with the marker coordinates. Then the ICP ~\cite{censi2008icp} post-processing is applied to optimize the coordinate alignment gap. Specifically, ICP optimizes the relative transformation between the partial point cloud obtained from the depth camera and the complete point cloud sampled from the reconstructed mesh.

\subsection{Expert Data Collection}

\label{sec:method:data-generation}
To verify the effectiveness of the real-to-sim-to-real system, following the designed task scenarios such as pick-and-place, we collect large-scale demonstrations using the script policy. 
During each rollout, we introduced domain randomization, including source/target object randomization, and robot arm's base position randomization.
% After reconstructing the whole scene in the simulator, we can use \our to generate high-quality simulation data automatically. 
% Following the above steps, objects and the robot are positioned and aligned under different task-specific requirements in the simulator, and both are randomly initialized in different poses for diversity. 
% A long-horizon task can be further divided into multiple stages based on the sequence of object interactions.  
Define an expert policy $\pi_\text{priv}(a_t | o_{\text{priv}, t})$ based on privileged information $o_{\text{priv}, t}$, such as the exact pose of target objects. We can then make use of $\pi_\text{priv}$ to interact with the simulator and generate observation-action pairs $(o_t, a_t)$.
Here, $o_t$ represents observations that are accessible from the real world, including images and proprioception data.
Specifically, we compute the 6D pose of the gripper at some key steps and use the motion planner based on the RRTConnect~\cite{rrtconnect} algorithm to plan the path between the key steps in the joint space.
During data generation, reject sampling is applied to filter out failed rollouts, improving the dataset quality and ensuring reliability.
We can continue to generate such data with the amount according to our needs, obtaining the simulated datasets $\mathcal{D}$. 

\section{Experiment}
\label{sec:experiment}
We mainly investigate the following three research questions based on the proposed \our system:
\begin{enumerate}[label=\textbf{Q\arabic*}, leftmargin=0.3in, itemsep=0pt, parsep=0pt, topsep=0pt]
    \item Can \our generate high-fidelity simulation data with small 3D sim-to-real gaps to benefit real-world manipulation problems?
    \label{ques:2}
    \item How does large-scale simulation data help in more challenging real-world tasks?
    \label{ques:3}
    \item Can \our rapidly construct simulated scenes and synthesize data at low cost?
    \label{ques:4}
\end{enumerate}

% \vspace{0.1in}
\subsection{Experimental Setups}

\noindent\textbf{Hardware platform.}
In this paper, we evaluate \our with a Franka Research 3 robot equipped with a parallel gripper. 
Two Realsense D435i depth cameras capture visual observations—one mounted on the end-effector and another positioned beside the robot for a third-person view.

\begin{figure*}[t]
    \centering
    % \vspace{-8pt}
    \includegraphics[width=.9\linewidth]{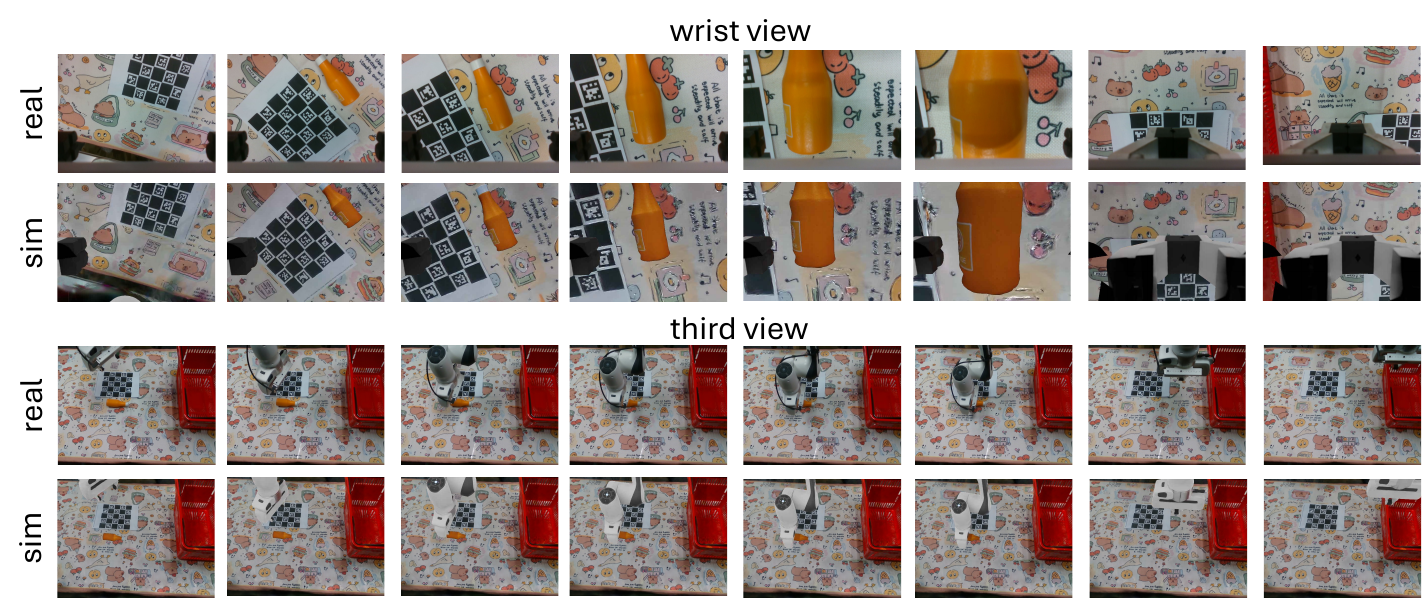}
    % \vspace{-8pt}  % Adjust this value as needed
    \caption{\textbf{Visual comparison between real and simulation.} Rendering results from our hybrid rendering method compared with photos captured by real-world cameras, highlighting the high fidelity and realism achieved by our approach.}
    \vspace{-8pt}
    \label{fig:visual qualitative comparison}
\end{figure*}

\noindent\textbf{Policy model and training details.}
% We train a policy model $\pi(\tau_{a, t} | o_t)$ on the generated simulation demonstration dataset $\mathcal{D}$.
% During our sim-to-real experiments, we adopt an ACT-structured policy~\cite{zhao2023learning} with DINOV2~\cite{oquab2023dinov2}.
% The images in the dataset are compressed using JPEG format.
% During training, we apply visual augmentations, and for evaluation, we use the temporal ensemble~\cite{zhao2023learning} technique.
% Appendix provides further details. 

During our sim-to-real experiments, we adopt an ACT-structured policy~\cite{zhao2023learning} with DINOV2~\cite{oquab2023dinov2}.
The images in the dataset are compressed using JPEG format.
During training, we apply visual augmentations, and for evaluation, the temporal ensemble~\cite{zhao2023learning} technique is used to smooth the final action for rollouts.

\label{sec:experiment_reconstruction_results}

\subsection{Zero-Shot Sim-to-Real Experiments}

\label{sec:experment_sim2real_results}

\ref{ques:2} is the main question that motivates us to build \our, thus we design three table-top manipulation tasks to investigate the usage of \our:\\
1.\texttt{Pick and drop a bottle into the basket.} The robot must pick an orange bottle randomly placed within a $25\text{cm} \times 35\text{cm}$ area on a table and place it into a basket. \\
% approximately 700 trajectories in the simulator.
% We collect about 1300 trajectories in the simulator.
2.\texttt{Place a vegetable on the board.}
A cucumber model is placed on the table in an area of $35\text{cm} \times 50\text{cm}$, and the cutting board's position is varied within a $35\text{cm} \times 80\text{cm}$ area. The robot must grasp the vegetable and accurately place it on the cutting board. \\
% We collected about 4000 trajectories for this task.
3.\texttt{Stack blocks.} The robot is tasked with stacking three black cubes on the table. Each cube is placed randomly within a $30\text{cm} \times 10\text{cm}$ area separately, and the robot must grasp and stack them.

% Tasks are rebuilt in simulation using \our, with 100 trajectories collected to train independent policies.
% Details of the rule-based policy and simulator evaluation implementation are in Appendix.
Tasks are rebuilt in simulation using \our, with 100 trajectories collected to train independent policies. As a baseline, we also adopt a rule-based policy that prioritizes picking objects closer to the robot base to reduce action variability.

% Details of the rule-based policy and simulator evaluation implementation are in Appendix~\ref{sub:task_details}.

\noindent\textbf{Visual quality evaluation.} An open-loop visual comparison is conducted by selecting a trajectory from the simulation dataset, replayed in the real world to capture ground-truth images with calibrated cameras. Visual similarity is evaluated from wrist and third-person perspectives, with qualitative results in Figure~\ref{fig:visual qualitative comparison}. These confirm that \our achieves high-quality visuals through accurate reconstruction and precise alignment. 

\begin{figure*}[ht]
    \centering
    \includegraphics[width=.9\textwidth]{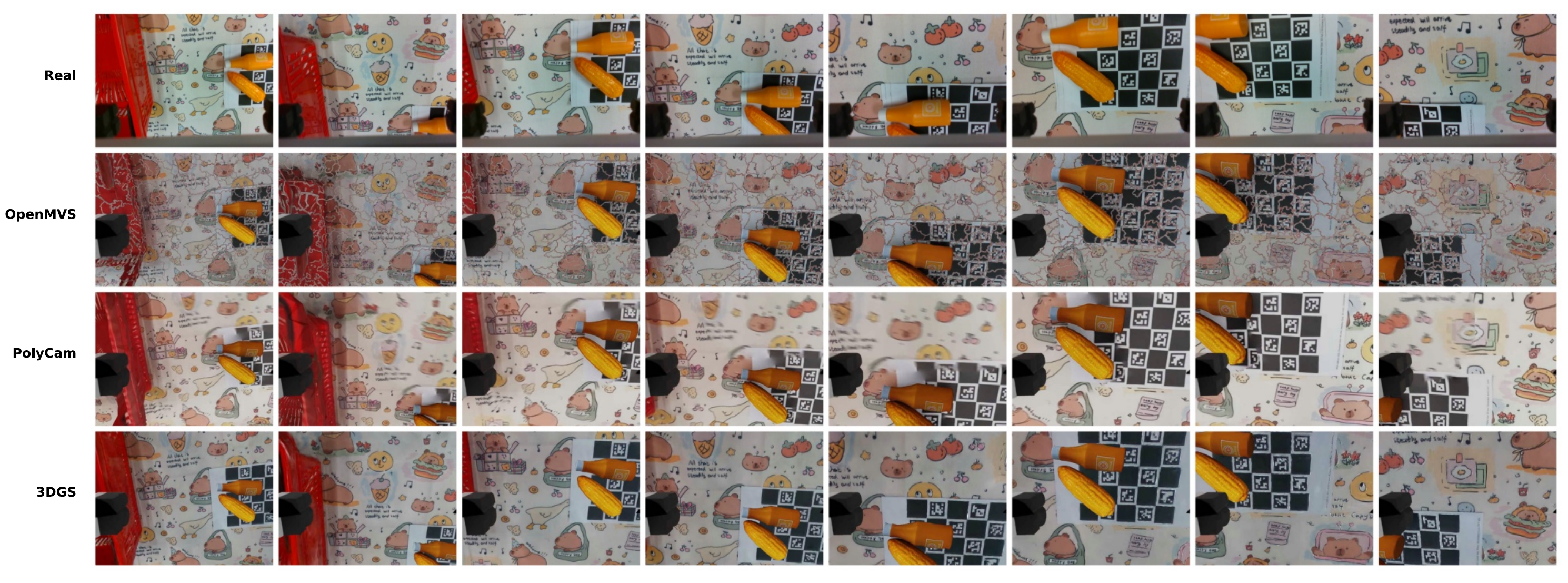}
    \caption{\textbf{Visual comparison of rendering approaches.} Rendering results of reconstruction outputs from PolyCam, OpenMVS, and 3DGS, compared with real-world photos.}
    \label{fig:compare_reconstrcution_methods}
    \vspace{-10pt}
\end{figure*}

We also compare with other reconstruction methods.
Visual comparisons of reconstruction results from more methods are shown in Figure ~\ref{fig:compare_reconstrcution_methods}. PolyCam tends to blur details in certain areas, OpenMVS's reconstructed mesh texture shows cracks, while 3DGS produces high-quality renderings. 
Table~\ref{tab:quantitative comparison} reports PSNR and SSIM for the visual sim-to-real gap, comparing \our with OpenMVS and PolyCam~\cite{ritotorne2024rialto}. Alignment error still causes minor pixel-level discrepancies, leading to a absolute lower PSNR and SSIM value in texture-rich scenes.
3DGS in \our surpasses PolyCam in both metrics and achieves higher SSIM than OpenMVS, despite comparable PSNR.
OpenMVS's PSNR is slightly higher than 3DGS's, likely because the background and crack colors are similar, making PSNR less sensitive. However, the cracks cause increased variance within patches, leading to a lower SSIM compared to 3DGS. These results validate \our's rendering design for minimizing visual gaps.

% \noindent\textbf{Sim-to-real results.}The real-world success rates of policies trained on simulation data from \our are evaluated, compared with a real-to-sim baseline, RialTo~\cite{ritotorne2024rialto}, and policies trained on 50 real-world trajectories. Doubling the simulation data yields performance comparable to real-world data, likely due to a minimal sim-to-real gap and differences in data collection methods, as detailed in Appendix. For comparison, AnyGrasp~\cite{fang2023anygrasp}, a state-of-the-art grasping method, is implemented with scripted primitives for the \texttt{pick and drop a bottle into a basket} task. AnyGrasp predicts only grasp poses, limiting its applicability to other tasks with non-fixed placement locations.

% \vspace{0.1in}
\noindent\textbf{Sim-to-real results.}The real-world success rates of policies trained on simulation data from \our are evaluated, compared with a real-to-sim baseline, RialTo~\cite{ritotorne2024rialto}, and policies trained on 50 real-world trajectories. Doubling the simulation data yields performance comparable to real-world data, indicating a minimal sim-to-real gap arising from differences in scene initialization and trajectory collection. For comparison, AnyGrasp~\cite{fang2023anygrasp}, a state-of-the-art grasping method, is implemented with scripted primitives for the \texttt{pick and drop a bottle into a basket} task. AnyGrasp predicts only grasp poses, limiting its applicability to other tasks with non-fixed placement locations.

The numerical results listed in Table~\ref{table:main_experiment} demonstrate that the data generated by \our help the policy achieve zero-shot sim-to-real transfer, showing strong performance even compared with the existing real-to-sim and state-of-the-art grasping method.
Furthermore, policies trained on our extensive synthetic dataset can achieve comparable or even slightly better performance than those only trained on real-world data, indicating the effectiveness and huge potential of high-fidelity simulation data.

%%%%%%%%%%%%%%%%%%%%%%%%%%%%%%
\subsection{Large-Scale Sim-to-Real Experiments}

\label{sec:large-scale-experiment}
To push the limit of utilizing synthetic data for real-world manipulation problems by answering \ref{ques:3}, we choose a \texttt{clear objects on the table} task: the robot has to clean up the table by picking all items that are randomly placed on the table in a $40\text{cm} \times 50\text{cm}$ area, and put them into the basket. This is a long-horizon task requiring the policy to sequentially pick both seen and unseen objects one by one. To provide a sufficient evaluation, we design four different setups:
\begin{itemize}[leftmargin=0.1in, itemsep=0pt, parsep=0pt, topsep=0pt]
    \item \noindent\textit{Seen}: four seen objects (the bottle, cucumber, corn, and eggplant) included in the data for training. 
    \item \noindent\textit{Unseen}: four objects (the green pepper, banana, red bowl, and momordica charantia) that are unseen during training. 
    \item \noindent\textit{Cluttered}: all eight objects from \textit{seen} and \textit{unseen}. 
    \item \noindent\textit{Darkness}: the testing brightness is significantly lower than that of the simulation data. 
\end{itemize}

\begin{figure}[tb]
    \centering
    % \subfigure[Real-world evaluation and robustness test for large-scale sim-to-real.]{
    \subfigure[Real-world test for large-scale sim-to-real.]{
        \label{fig:scale_results}
        \includegraphics[width=.95\linewidth]{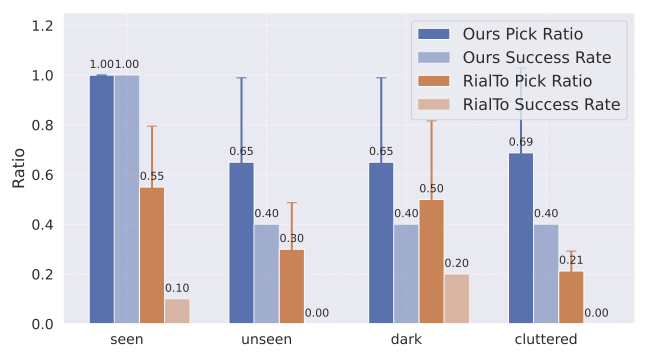}  
    }
    \subfigure[Data scaling effects.]{
        \centering
        \includegraphics[width=.95\linewidth]{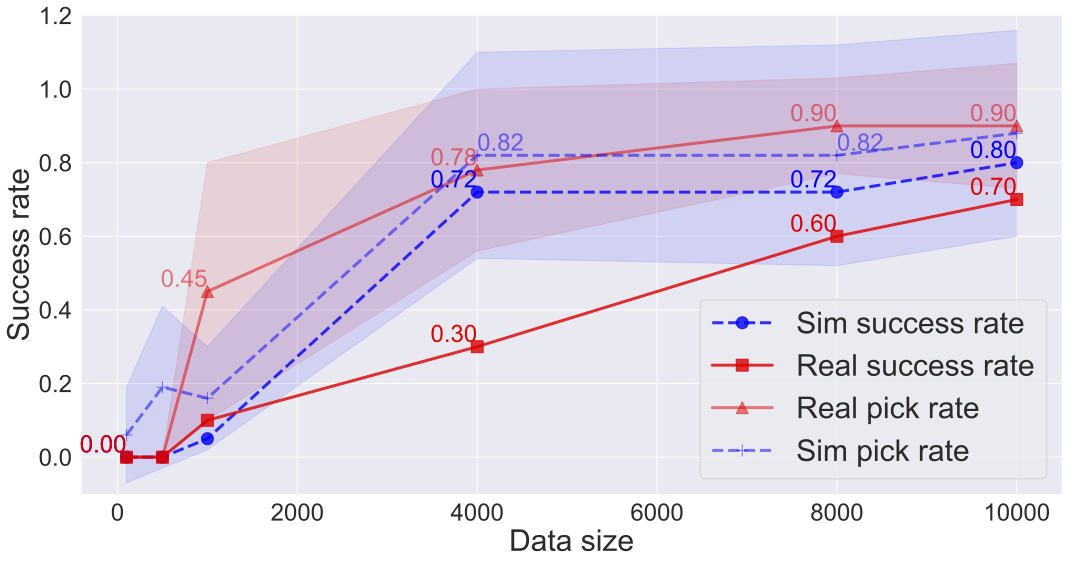}
        \label{fig:dataset-size}    
    }
    \vspace{-8pt}
    \caption{(a) \textit{Success rate} denotes the proportion of trials where all objects were grasped, while \textit{pick rate} represents the proportion of objects grasped relative to the total on the table. (b) The success rate and pick rate were tested on seen objects in the real world. See qualitative results in the \href{https://re3sim.github.io/}{ website}.}
    \vspace{-10pt}
\end{figure}

All models are evaluated on four \textit{seen} objects. The results are presented in Fig.~\ref{fig:scale_results}. Each setup was tested 10 times, with two additional automatic grasp attempts per object in case of failure.

\begin{table}[ht]
    % \vspace{-8pt}
    \caption{\textbf{Background rendering methods comparison.} Quantitative results of the reconstruction result in terms of PSNR and SSIM. Results are computed from 1029 robot arm positions in the wrist view.}
    % Our approach (\our) using 3DGS outperforms Polycam (mesh-based method) as used in \citet{ritotorne2024rialto}.

    % during a single grasping task
    
    \label{tab:quantitative comparison}
    \centering\small
    \begin{tabular}{ccccc}
    \toprule
        Background Rendering & PSNR & SSIM  \\ \midrule
        Polycam & \makecell{$11.52 $ $\pm 1.40$} & \makecell{$0.34$ $\pm 0.04$}  \\
        OpenMVS & \makecell{$\boldsymbol{13.40}$$ \boldsymbol{\pm 0.96}$} & \makecell{$0.27$ $\pm 0.03$} \\
        3DGS & \makecell{$\boldsymbol{13.29}$ $\pm \boldsymbol{1.11}$} & \makecell{$\boldsymbol{0.37}$ $\boldsymbol{\pm 0.04}$} \\
        \bottomrule
    \end{tabular}
    \vspace{-15pt}
\end{table}

\begin{table*}[t]
    % \small
    \vspace{5pt}
    \caption{\textbf{Quantitative results of real-to-sim-to-real evaluation.}Real-world manipulation tasks are tested 20 times each, without retries on failure. Policies are trained with 100 simulation episodes (RialTo+IL, \our+IL) or 50 real-world episodes (Real+IL). AnyGrasp with scripted primitives (AnyGrasp+Prim) is evaluated in the scenario with fixed placement locations.}
    \label{table:main_experiment}
    \centering
    \begin{tabular}{lccccccc}
        \toprule
        Real-world Tasks  & \makecell{RialTo+IL} & \makecell{AnyGrasp+Prim.}  & \makecell{Real+IL} & \makecell{\our+IL} 
        % & \makecell{\{Real+\our\}+IL}  
        \\
        \midrule
        \multirow{1}{*}{Pick and drop a bottle into the basket} & 0.4 & 0.9 & \textbf{0.8} & 0.75 & \\
         % & Diffusion &  \\ \midrule
        % \multirow{1}{*}{Pick up two objects to basket} & - & 0.7 & 0.6 & 0.7 & \\
        % \multirow{1}{*}{Pick up two objects to basket} & - & 0.2 &  & 0.4 & \\
        % & Diffusion &\\ \midrule
        \multirow{1}{*}{Stack cubes} & 0 & - & 0.15 & \textbf{0.25} & \\
        % & Diffusion \\ 
        \multirow{1}{*}{Place a vegetable on board} & 0.45 & - & 0.6 & \textbf{0.75}\\
        \bottomrule
    \end{tabular}
    \vspace{-10pt}
\end{table*}

% \begin{figure}[tb]
%     \centering
%     \includegraphics[width=1.\linewidth]{images/scale_results.pdf}
%     \vspace{-8pt}
%     \caption{\textbf{Real-world evaluation and robustness test for large-scale sim-to-real.} The \textit{success rate} reflects the proportion of trials in which all objects were successfully grasped, while the \textit{grasp rate} indicates the proportion of objects grasped relative to the total number on the table. See qualitative results in the \href{http://xshenhan.github.io/Re3Sim/}{ website}.}
%     \label{fig:scale_results}
% \end{figure}

% \begin{figure}[tb]
%     \centering
%     \includegraphics[width=1.\linewidth]{images/dataset_size_ablation.pdf}
%     \caption{\textbf{Data scaling effects}, tested on seen objects in the real world. }
%     \label{fig:dataset-size}
% \end{figure}

% \begin{figure}[tb]
%     \centering
%     \includegraphics[width=1.\linewidth]{images/scale_results.pdf}
%     \vspace{-8pt}
%     \caption{\textbf{Real-world evaluation and robustness test for large-scale sim-to-real.} The \textit{success rate} reflects the proportion of trials in which all objects were successfully grasped, while the \textit{grasp rate} indicates the proportion of objects grasped relative to the total number on the table. See qualitative results in the \href{http://xshenhan.github.io/Re3Sim/}{ website}.}
%     \label{fig:scale_results}
% \end{figure}

% \begin{figure}[tb]
%     \centering
%     \includegraphics[width=1.\linewidth]{images/dataset_size_ablation.pdf}
%     \caption{\textbf{Data scaling effects}, tested on seen objects in the real world. }
%     \label{fig:dataset-size}
% \end{figure}
% \vspace{0.1in}
\noindent\textbf{Result analysis.}
Despite training on only five objects, the policy generalizes effectively to grasping novel objects of similar size, regardless of shape or color variations. We hypothesize that the robot exploits the static scene, detecting objects via background color contrasts. Additionally, objects with varying shapes often have similar grasping positions, enabling the robot to execute successful grasps.  Larger datasets with multiple objects enhance the policy's success rate, whereas smaller datasets limit its performance. Compared with RialTo-IL~\cite{ritotorne2024rialto}, policies trained on data from \our exhibit superior performance across all settings. Moreover, the policy shows robust performance under varying lighting conditions.

\begin{table}[tb]
    \caption{\textbf{Human effort in reconstruction.} Estimated scene reconstruction times at the table level, alongside human effort for object reconstruction using ARCode.}
    \label{tab:human-effort}
    \centering
    \begin{tabular}{cccc}
    \toprule
    Input Types & Video & Images & ARCode \\
    \midrule
    Human Efforts (s) & 51.5 & 84.5 & 60.5 \\
    \bottomrule
    \end{tabular}
    \vspace{-10pt}
\end{table}

\begin{table}[tb]
    \caption{\textbf{Time cost for simulation data collection.} Time to collect 100 simulation episodes per task using eight RTX 4090 GPUs.}
    \label{tab:data-collection-time}
    \centering
    \begin{tabular}{lccc}
    \toprule
        Tasks & Time Cost (minutes) \\
    \midrule 
    Pick and drop a bottle into the basket  & 12.35 \\
    Place a vegetable on the board & 13.78\\
    Stack blocks & 6.45 \\
    \bottomrule 
    \end{tabular}
\vskip -0.1in
\end{table}

\begin{table}[h]
    \caption{\textbf{Average time consumption per step.} Breakdown of time consumed by each process. 
    % Gaussian splatting rendering introduces low additional computational overhead, with the entire process taking an average of 41.46 ms per step.
    }
    \label{tab:time_consumed}
    \centering
    \begin{tabular}{lc}
    \toprule
         Process & Time (ms) \\
    \midrule
         Physics Simulating & 26.64\\
         Gaussian Splatting Rendering & 12.93\\
         Motion Planning& 0.36\\
         Others & 1.53\\
         Total Time & 41.46 \\
    \bottomrule
    \end{tabular}
    \vspace{-0.2in}
\end{table}

We further explored how the success rate in simulation and real-world settings changes with varying amounts of data. As shown in Fig.~\ref{fig:dataset-size}, increasing the size of the synthetic dataset generated by \our leads to a significant improvement in imitation learning performance. When the amount of simulation data is limited, the real-world performance is almost negligible. Doubling the data size often results in a significant improvement in success rate until convergence at a high performance. 

\subsection{Sim-Real Consistency}

% \begin{figure}[!htbp]
%     \centering
%     \includegraphics[width=0.9\linewidth]{images/consistency.pdf}
%     \caption{\textbf{Analysis of real and simulated evaluation consistency.} Models that perform better in simulation tend to perform better in real-world settings as well.}
%     \label{fig:sim policy consistency}
% \end{figure}

We evaluated the consistency of policy performance between simulated and real-world environments using policy ACT. In this experiment, we evaluate policies trained at different stages (from 0 to 120000 training steps) and with 3 different seeds on the \texttt{pick and drop a bottle into a basket} task.

As illustrated in Fig.~\ref{fig:teaser} (d), the Pearson correlation coefficient of \our is 0.924, while that of RialTo is 0.710, indicating the correlation between performance in simulation and the real world of \our is better. Specifically, models that achieve higher success rates in simulation tend to exhibit similar success rates in real-world testing.
When the success rate in simulation is low, it is often difficult for the model to perform well in real-world scenarios. This could be due to alignment errors, leading to a small sim-to-real gap, or the fact that the number of real-world test instances is much lower than in simulation, potentially introducing bias.
This consistency suggests that policies trained with \our are well-suited for real-world deployment, reflecting a minimal sim-to-real gap.

\subsection{System Efficiency}

To validate \ref{ques:4}, we record and compute the time of \our to reconstruct both the background and objects, along with the time of generating demonstrations.

% ARcode 55s 66s

\begin{table}[h!]
    % \small
    \vspace{5pt}
    \caption{\textbf{Ablation on foreground rendering.} Task success rates for 3DGS vs. ARCode rendering.}
    \label{tab:abla-gs-fore}
    \centering
    \begin{tabular}{lcc}
        \toprule
Real-world Tasks & \makecell{Pick and drop a bottle\\into the basket} & \makecell{Place a vegetable\\on board} \\
        \midrule
        3DGS   & 0.70 & \textbf{0.77} \\
        ARCode & \textbf{0.75} & 0.75 \\
        \bottomrule
    \end{tabular}
    \vspace{-8pt}
\end{table}

% \vspace{0.1in}
\noindent\textbf{Human effort during reconstruction.} In \our, background reconstruction only requires a set of images or a short video, plus a single depth-aligned image. While the process is straightforward, it still involves some human effort. We report average reconstruction times for both foreground and background, based on five users with basic knowledge of modern phones or cameras. The workflow is simple: (1) capture photos or video using a phone or camera, and (2) follow ARCode's guided interface. Users typically need just 1–3 practice runs to become proficient. As shown in Table~\ref{tab:human-effort}, video input saves time but may slightly reduce quality due to motion blur, while images yield better results with slightly more effort.

\noindent\textbf{Data collection cost.}
Table~\ref{tab:data-collection-time} shows the time required to collect 100 simulation episodes for each task in Section~\ref{sec:experment_sim2real_results}, using 8 RTX 4090 GPUs.
The time required for data collection is much lower than the time needed for teleoperation in real-world scenarios, especially considering \our only involves machine runtime, and the latter mostly requires the time of data collection experts. This demonstrates our capability to efficiently generate large-scale simulation data at minimal cost. The breakdown of the time consumption of each step is detailed in Tab.~\ref{tab:time_consumed}.
% ~\ref{sub:experiment_details}

\noindent\textbf{Time consumption breakdown.} We measured the average time spent on each phase during the data collection of the \textit{pick and drop a bottle} task in the simulator over 60 steps, as shown in Tab.~\ref{tab:time_consumed}. 
The reconstructed mesh, with numerous vertices and faces, demands more time for physics simulation. 
Rendering 480p images from two camera views using 3DGS adds 12.93 milliseconds, representing one-third of the total time per step. \our operates at approximately 24 frames per second (FPS) with two cameras, ensuring real-time performance and visual fidelity.

\subsection{Impact of Mesh-Based vs. 3DGS Rendering Methods}

We explore using 3DGS for foreground rendering as an alternative to ARCode-reconstructed meshes. Leveraging the GSDF~\cite{yu2024gsdf} framework, we generate aligned mesh–3DGS pairs for integration into our hybrid system. As reported in Table~\ref{tab:abla-gs-fore}, substituting ARCode with 3DGS offers no appreciable benefit in task success. In light of this, we continue to use ARCode-based meshes as the default in our main experiments.

% \noindent\textbf{Multi-view 3D Reconstruction.}
% \todo{xiaoyang}

% We do not need this
% \noindent\textbf{Manipulation policy.}
% Recently, a series of algorithms have achieved impressive results in robotics manipulation tasks. Algorithms such as ACT~\cite{zhao2023learning}, Diffusion Policy~\cite{chi2023diffusion}, RVT~\cite{goyal2023rvt}, and 3D Diffusion Policy~\cite{ze20243d} have demonstrated strong performance on in-domain tasks even when trained on small-scale real-world datasets. Simultaneously, algorithms~\cite{li2023vision, brohan2022rt,o2023open, kim2024openvla,wu2023unleashing,cheang2024gr2generativevideolanguageactionmodel, tian2024seer} leverage pre-trained datasets to train models that exhibit generalization across multiple tasks and diverse scenarios. Meanwhile, many algorithms utilizing pre-trained vision language models to finish robotics tasks, such as R3M~\cite{nairr3m}, MVP~\cite{xiao2022mvp,radosavovic2023mvp}, VIP~\cite{ma2023vip}, and VC-1~\cite{majumdar2023vc1} pre-trained the vision models on ego-centric dataset~\cite{grauman2022ego4d}. 

% object-centric human motion dataset
% robot dataset

\vspace{0.1in}
\section{Related Work}

\noindent\textbf{Sim-to-real.}
Sim-to-real transfer requires techniques to enable policies to successfully adapt from simulation to the real world. The most direct approach is improving simulators \cite{todorov2012mujoco, isaacsim}, which reduces the sim-to-real gap. Other methods, such as domain randomization~\cite{mehta2020active, chen2021understanding} and system identification~\cite{ song2024systemid, pmlr-v100-allevato20a}, also aim to bridge this gap.

\noindent\textbf{Real-to-sim-to-real.}
Many recent works leverage real-world data to enhance simulation models. Reconstruction methods integrated with grasping techniques, such as Evo-NeRF~\cite{kerr2023evo} and LERF-TOGO~\cite{lerftogo2023}, enable the grasping of objects using only RGB images. 
% Prior approaches, such as GraspNerf~\cite{dai2023graspnerf}, introduced a generalizable NeRF that achieves real-time grasping. Other works, including Evo-NeRF~\cite{kerr2023evo} and LERF-TOGO~\cite{lerftogo2023}, utilize depth images rendered by NeRF to generate grasping poses.
% Specifically, LERF-TOGO leverages depth information from multiple views to create a dense point cloud, which is subsequently processed by GraspNet~\cite{fang2020graspnet}.
% Recently, 3D Gaussian splatting~\cite{kerbl3Dgaussians} has gained significant attention in robotics due to its fast rendering speed and explicit representation. GaussianGrasper~\cite{zheng2024gaussiangrasper} uses 3DGS for scene reconstruction and normal-guided grasp generation. Similarly, SplatMover~\cite{shorinwa2024splat} introduces a grasp-splat module integrating affordance and semantics within 3DGS. GraspSplats~\cite{ji2024graspsplats} replaces object representation in earlier grasping networks with 3DGS.
Meanwhile, GaussianGrasper~\cite{zheng2024gaussiangrasper} and GraspSplats~\cite{ji2024graspsplats}, use 3D Gaussian splatting~\cite{kerbl3Dgaussians} for fast rendering and explicit representation in robotics tasks. Additionally, methods like URDFormer~\cite{urdformer} and Articulate Anything~\cite{articulateanythingle2024} use a single image to reconstruct the environment directly, allowing for collecting large amounts of data for imitation learning or reinforcement learning. These approaches enhance data by varying articulations and leveraging various articulations from simulation datasets to train models deployable in the real world.

% Another approach involves constructing a simulator with a smaller sim-to-real gap using real2sim methods. This allows for the collection of large amounts of data in the simulator for imitation learning or the use of reinforcement learning to interact with the environment, training an end-to-end model rather than using a scripted policy as the actuator. Specifically, URDFormer~\cite{urdformer}, Digital Cousins~\cite{acdcdai2024}, and Articulate Anything~\cite{articulateanythingle2024} use a single image to reconstruct the environment. 
% URDFormer utilizes URDF as a representation, employing generative models to create a photo-URDF dataset, thereby training models to generate URDFs from images. Digital Cousin and Articulate-Anything retrieve similar meshes from datasets to construct scenes in simulation.
% They enhance their data by using different articulations of similar sizes from simulation datasets, and by collecting a large amount of data similar to real environments in geometry and semantics, they train models that are able to be deployed in the real world. 

With the help of 3DGS works like RoboStudio~\cite{robostudio}, SplatSim~\cite{qureshi2024splatsim}
and RoboGSim~\cite{li2024robogsim} use multi-view images or video for world reconstruction, improving multi-view rendering quality with minimal cost. These methods are especially effective in manipulation tasks involving multiple objects or occlusions. 1) RoboStudio focuses on reconstructing the URDF of a robot, offering a photorealistic rendering result and an accurate collision mesh.
2) SplatSim utilizes pre-obtained 3D models of objects and backgrounds to collect trajectories in the physics simulator and then re-render them with 3DGS to reduce the visual gap between simulated and real-world environments, but acquiring the 3D models is hard for many tasks. 
3) RoboGSim compares rendering quality, validates high-quality rendering results for novel pose synthesis, and shows the potential for evaluating various manipulation algorithms. However, its sim-to-real validation remains limited.
Different from them, \our reconstructs both geometric and visual aspects with small gaps, and validates robot policies trained on simulated data through extensive experiments in the real world.

\section{Conclusion}

\label{sec:conclusion}

We introduced \our, a novel Real-to-Sim-to-Real system that integrates Gaussian splatting with NVIDIA Isaac Sim's PhysX engine, improving scene reconstruction and sim-to-real transfer for robotic manipulation tasks. Extensive experiments demonstrate its efficacy and scalability, showing that \our significantly reduces the sim-to-real gap and enables reliable real-world task performance. Models trained on large simulated datasets achieved competitive success rates compared with those trained on real-world data, highlighting the potential \our to generate diverse, high-quality data for pre-training large-scale robot models.
% Our sim-to-real transfer experiments showed that \our significantly reduces the sim-to-real gap, enabling trained policy models to perform reliably in real-world scenarios. Models trained on extensive simulated datasets achieved competitive success rates, often surpassing mesh-based rendering methods such as IL-RialTo and matching the performance of models trained with real-world data and the AnyGrasp-Rule method. These results highlight \our's ability to generate high-quality and diverse data essential for training robust imitation learning models.

 % and advancing robotic applications.

% \noindent\textbf{Limitations.} 
 % \our is currently limited to rigid objects and has not yet been extended to reconstruct articulated, deformable objects, or liquids. Additionally, it relies on manually defined physics parameters rather than system identification methods. Rule-based policies for data collection become challenging as task complexity increases. We leave addressing these limitations to future work. 

% \vspace{0.1in}
\noindent\textbf{Limitations.} Currently, \our does not support the reconstruction of deformable objects or liquids. The system also relies on manually defined physics parameters rather than automated system identification. Furthermore, rule-based policies for data collection become increasingly impractical as task complexity grows. Addressing these limitations is left to future work.

\section*{Acknowledgment}
The SJTU team is supported by National Natural Science Foundation of China (62322603).

% Our method is currently limited to rigid objects and cannot reconstruct articulated, deformable objects, or liquids. Furthermore, we do not utilize system identification methods, instead relying on default values for physics parameters
% , which are inaccurate for objects with unique physical characteristics such as irregular centers of mass. 
% Additionally, rule-based policies for data collection become challenging as task complexity increases. 
% Future work could focus on developing a unified pipeline for high-fidelity articulation reconstruction and exploring more user-friendly data generation methods to address these limitations.

% \section*{Impact Statement}

% The proposed \our system provides an intriguing way of generating simulated robot demonstrations for real-world tasks. This work is meant to release the human burden of collecting large-scale real-world data, improve efficiency, and push the limit of generalist robot policies with data scaling law. \our can be applied across various robot applications and thus may cause potential unemployment issues.
% \newpage
% \begin{thebibliography}{99}
\bibliographystyle{IEEEtran}
\bibliography{main}  % .bib
% \end{thebibliography}

\end{document}